\documentclass[letterpaper, 10 pt, journal, twoside]{ieeeconf}

\IEEEoverridecommandlockouts  

\usepackage{todonotes}

\usepackage{verbatim} 

\usepackage{graphicx}
\usepackage{subcaption}

\usepackage{tabulary}
\usepackage{tabularx}
\usepackage{mathtools}
\usepackage{textcomp}
\usepackage{multirow}
\usepackage{amsmath,bm}
\usepackage{amssymb}  
\usepackage{url}
\usepackage{algorithm}
\usepackage{algpseudocode}
\usepackage{color}
\usepackage{ctable}
\usepackage{placeins}

\newcommand{\dorian}[1]{{\color{blue}dt: #1}}

\def\R{\mathbb{R}}

\newcommand{\vecx}{\mathbf{x}}

\newcommand{\mean}{\mathbf{\mu}}

\newcommand{\matr}[1]{\mathbf{#1}}
\newcommand{\mySigma}{\mathbf{\Sigma}}

\title{\LARGE \bf 3D Move to See: Multi-perspective visual servoing for improving object views with semantic segmentation}

\author{Chris Lehnert$^{*}$, Dorian Tsai$^{*}$, Anders Eriksson$^{*}$ and Chris McCool$^{*}$%
\thanks{$^{*}$C. Lehnert, D. Tsai, A. Eriksson and C. McCool are with the Queensland University of Technology (QUT), Brisbane, Australia {\tt\small \{c.lehnert, dy.tsai, anders.eriksson, c.mccool\}@qut.edu.au}}
}

\begin{document}

\maketitle



\begin{abstract}
In this paper, we present a new approach to visual servoing for robotics, referred to as 3D Move to See (3DMTS), based on the principle of finding the next best view using a 3D camera array and a robotic manipulator to obtain multiple samples of the scene from different perspectives. The method uses semantic vision and an objective function applied to each perspective to sample a gradient representing the direction of the next best view.
The method is demonstrated within simulation and on a real robotic platform containing a custom 3D camera array for the challenging scenario of robotic harvesting in a highly occluded and unstructured environment.
It was shown on a real robotic platform that by moving the end effector using the gradient of an objective function leads to a locally optimal view of the object of interest, even amongst occlusions. The overall performance of the 3DMTS method obtained a mean increase in target size by $\bm{29.3\%}$ compared to a baseline method using a single RGB-D camera, which obtained $\bm{9.17\%}$. The results demonstrate qualitatively and quantitatively that the 3DMTS method performed better in most scenarios, and yielded three times the target size compared to the baseline method.
The increased target size in the final view will improve the detection of key features of the object of interest for further manipulation, such as grasping and harvesting.

\end{abstract}

\section{Introduction}

Moving a robot or autonomous system to better view a target object has usually been cast as a visual servoing task.
Visual servoing (VS) is a widely applicable robot control technique that directly uses visual information by placing the camera into the control loop. 
In particular, image-based VS tracks visual features such as points and lines to directly estimate the required rate of change of camera pose. 
Typically, image-based VS minimizes the error between the current image $I_c$ and a known image template $I_{t}$ that corresponds to the desired pose~\cite{hutchinson1996visualservotutorial,chaumette2006basicApproaches}. 
A strong assumption that is often made, and is critical to the success of such approaches, is that $I_{t}$ is known \textit{a~priori}. However, there are many practical cases where this is not feasible.

Knowing $I_{t}$ \textit{a~priori} is difficult in unstructured environments such as robotic harvesting. 
In robotic harvesting, there is a particular crop to be harvested, but it does not have a single image template $I_{t}$ due to the natural variation of the crop in terms of size and appearance.
Furthermore, as this is an unstructured environment there are other variations due to lighting, pose, orientation and also occlusion.
Therefore, even if there was a template image $I_{t}$ that well described the object of interest (fruit), the appearance of this object would still vary dramatically.
So much so that classical VS could no longer be applied.

\begin{figure}[!t]
  \begin{center}
  {
    \begin{subfigure}[t]{0.48\columnwidth}
        \includegraphics[ width=\columnwidth]{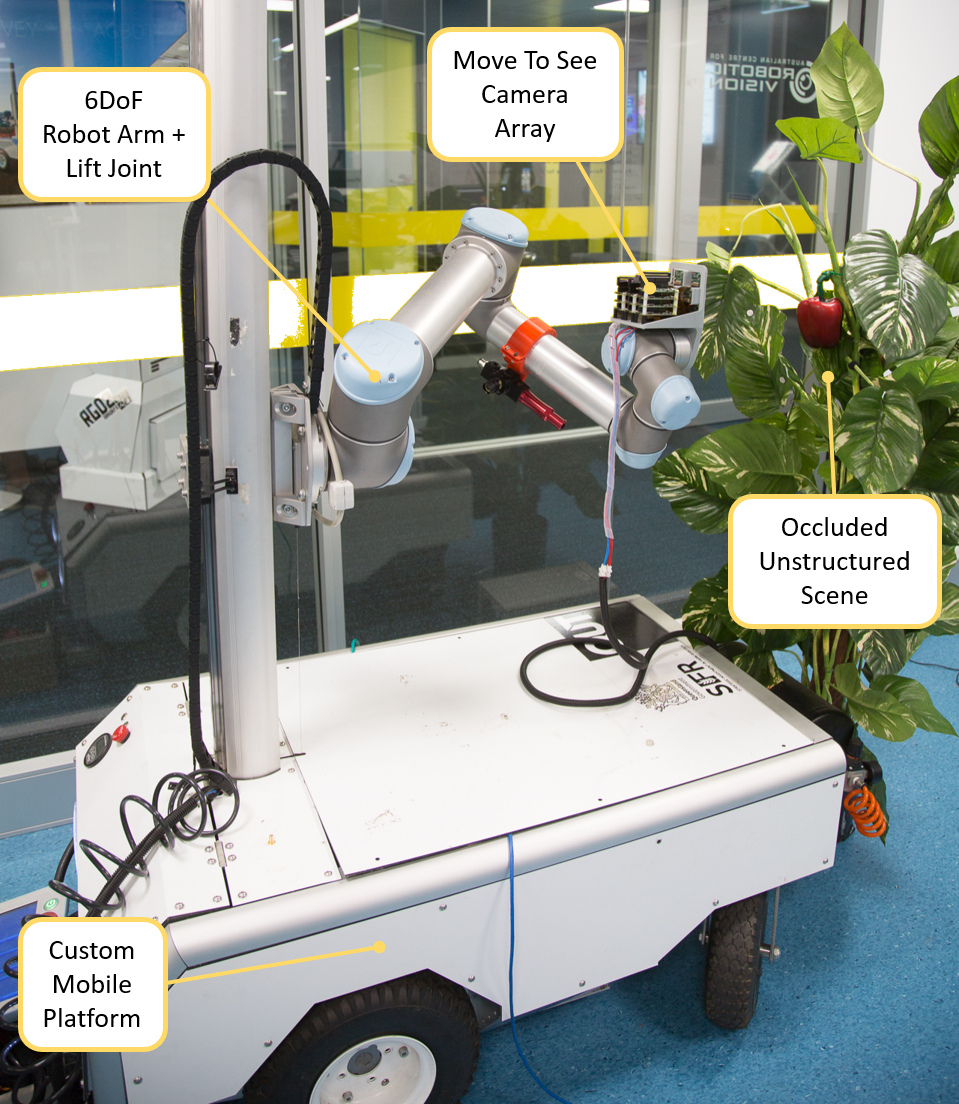}
        \caption{}
        \label{fig:harvey_move_to_see_platform}
    \end{subfigure}
    \begin{subfigure}[t]{0.48\columnwidth}
        \includegraphics[ width=\textwidth]{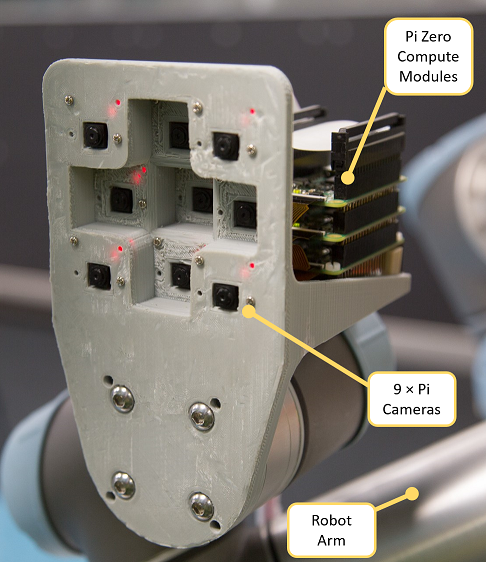}
        \caption{}
    \end{subfigure}
  }
  \end{center}
    \caption{(a) The robotic platform with the multi-perspective 3D camera array used to optimised the view point for occluded and unstructured scenes. (b) The prototype 3D camera array mounted on the robot arm. The camera contains nine pi cameras and ten pi zero computer modules.}
    \label{fig:FrontImage}
    \vspace{-20pt}
\end{figure}

Instead of performing VS, where a template $I_{t}$ is known in advance, we focus on an alternative problem setting: \textit{``how can we move the camera to better view the object of interest?''} often referred to as the next best view problem.
With this problem statement we can find the object of interest through a semantic segmentation of the scene, which has advanced considerably over the past few years~\cite{Shelhamer17_1,Jegou17_1:conference}.
If we could capture multiple views of the scene, the camera could then be moved in the locally optimal direction by choosing to move the camera towards the view where more of the object of interest (obtained through semantic segmentation) was observed.
A key problem with such an approach is how to obtain the multiple views of the scene in a realistic time frame; that is, without explicitly moving the camera to all of the required views.

In this paper we explore the above question and explore the potential to solve it using a camera array combined with semantic scene segmentation.
By using an array of cameras, we can instantaneously capture multiple similar views of the same scene from slightly different positions. 
Semantic segmentation of the target is then used to rate the current ``next best view'' from each camera in the array.
A score is assigned to each view and is used to form a gradient to move the robot to better view the target.
Constraints of the workspace and valid movements of the robotic arm are also introduced.
The robotic arm then moves in the desired direction and takes another scene, to continue servoing along the locally optimal movement.
Such an approach could one day be used for real-world robots such as Harvey the sweet pepper harvester (see Fig.~\ref{fig:FrontImage}), to automatically move around occlusions to the optimal viewing position for harvesting crops, which would increase robot harvesting accuracy~\cite{sa2017peduncle}.

Our main contributions are the following. First, we develop the theoretical formulation for the gradient ascent of the next best view using multiple perspectives from a 3D camera array. Second, we propose the design of a 3D camera array, which provides a range of perspectives to assist in calculating the target gradients. Finally, we validate our approach in simulation and on a real robotic platform and demonstrate that our approach improves object view for further manipulation in highly occluded and unstructured environments. 


\section{Related Work}

VS has been used in applications that require visual feedback, close-range manipulation, and fast action, from docking planetary rovers~\cite{tsai2013docking}, flying quadrotors~\cite{bourquardez2009ibvsQuad} to picking produce~\cite{Mehta2014,Baeten2008,Han2012}. Malis et al. extended classical VS techniques to multiple cameras
~\cite{malis2000multiCameraVS}. 
However, the number of cameras was simply used to increase the vision sample rate (more views, thus more features), and the impact of occlusions was not considered. 
Recently, VS with light field cameras using augmented image features demonstrated improved occlusion-handling performance under field-of-view constraints~\cite{tsai2016lfvisualservo}. The multiple views from the light field camera provided some redundancy against occlusions. However, the limited baseline and field-of-view of the custom mirror-based camera array limited the applicability of their system, and both of the above methods required a known $I_t$.
We visual servo with a feedback loop, but our method does not require an explicit $I_t$ known beforehand, and exploits the common geometry of our camera array to simultaneously simplify the control task and navigate occlusions in unstructured environments.


Feature-based VS relies on tracking and matching features, which can be prone to inaccurate correspondences. Direct VS methods avoid tracking and matching by directly using the image intensities by way of minimising the error between the current and desired image, but suffer from small convergence domains in comparison~\cite{collewet2011photometricVS}. To improve the convergence domain, Bateux et al. considered direct VS in the context of the ``next best view'' problem, using single camera with a particle filter to project $I_t$ to a desired pose. The error between the projected image and $I_t$ was used to automatically drive the robot towards the convergence area for conventional direct VS~\cite{Bateux15_1}.
Although the error was minimised between the current and next images, the poses projected by the particle filter were random, resulting in a path towards the goal pose that was not necessarily smooth or optimal with respect to the amount of physical motion required to reach the goal. Without an explicit $I_t$, our work uses a 3D camera array to produce an smooth trajectory towards the object.

Without a prior $I_t$, active vision techniques that actively control camera parameters, or camera motion, to remove ambiguities about a scene, can be used to automate robot tasks in unstructured environments~\cite{chen2011activeVisionSurvey}.
Zhang et al. combined VS with next best view theory to actively guide a robot and automatically reconstruct the surface of a 3D object~\cite{zhang2015visualServoingNextBestView}.
However their method did not consider occlusions. Maver et al. was the first to consider viewpoint selection driven by occlusions using a range finder~\cite{maver1993occlusionsPlanningNextView}, and
more recently, Sheinin et al. generalised the next best view concept to scattering media in a challenging underwater environment~\cite{sheinin2016next}. The poses of a light system and single camera were actively controlled to map scenes with large occlusions. However, their system was focused on mapping the environment to reduce pose uncertainty, and was limited to static scenes because they only used one camera. 
Using multiple cameras allows our method to handle more dynamic scenes. 
And because we jointly use a camera array to obtain multiple viewpoints, and an image segmentation/image-based approach within the objective function there is no need to explicitly estimate pose. 
This makes our approach much simpler, faster, and less prone to modelling and calibration errors.




\section{3D Move to See (3DMTS) Method}

In this paper we propose an alternative approach based on the idea of using multiple perspectives from a 3D camera array combined with semantic scene segmentation.
We propose to take multiple views of a scene at once using a 3D camera array, semantic segmentation of the object of interest is then used to rate the quality of each view in the array. 
This collection of individually rated views  
are then used to form a gradient (based on an objective function) to move the robot, in this case a robotic arm, to better view the object of interest.
Constraints, taking into account the workspace and valid movements of the robotic arm, are also introduced.
The robotic arm then moves in the desired direction and takes another image of the scene, to continue taking the locally optimum movement.
Such an approach could one day be used for real-world robots such as Harvey the sweet pepper harvester (see Fig.~\ref{fig:FrontImage}), to automatically move to the optimal viewing position.

\subsection{Objective Function}

The objective function should represent the goal of the system. For this work the main objective is to maximise the information about the detected object while accounting for the kinematic workspace of the robot. For this reason the objective function is the weighted sum of two criteria, the normalised size of the detected object in the scene, $p$ and a robot manipulability score, $m\bm{(q})$, of the given robot arm configuration variables, $\bm{q}$. The objective function that we wish to maximise is defined as
\begin{figure}
    \centering
    \includegraphics[ width=\columnwidth]{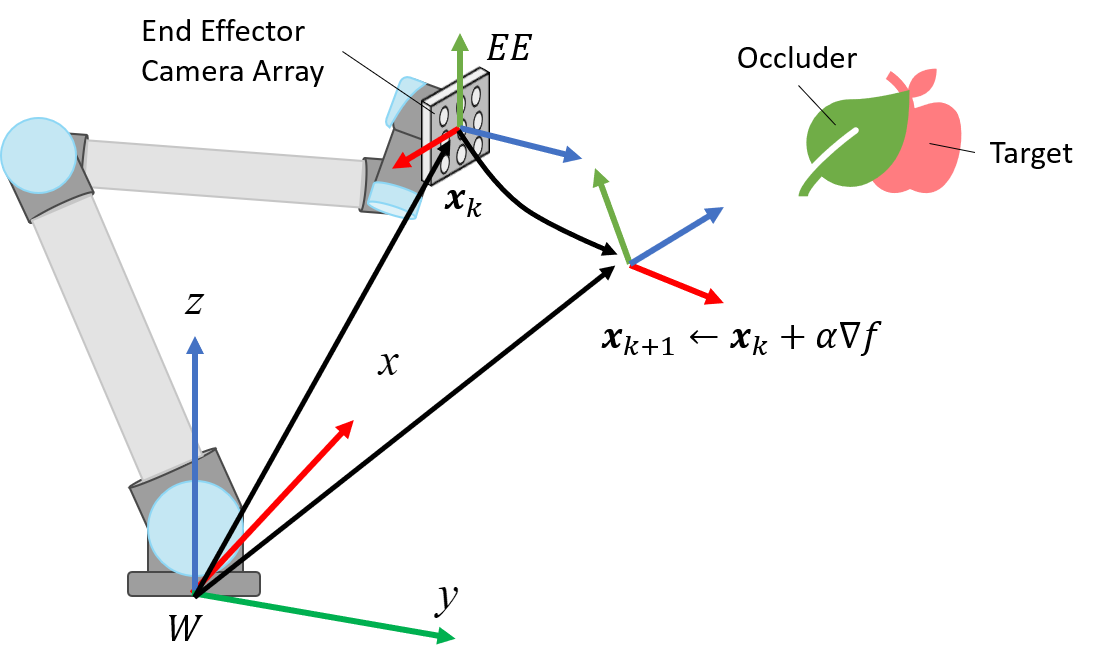}
    \caption{Diagram of the move to see concept. The camera array captures multiple measurements of the directional derivative to estimate the gradient, $\nabla f(\bm{x})$, at its current location $x_k$. Where $x_k$ is the translation vector between the world reference frame, $W$ and the end effector reference frame $EE$. The robot end effector is then moved to $x_{k+1}$ which is the direction of gradient ascent which maximises the objective function $f(x)$.}
    \label{fig:robot_diagram}
    \vspace{-1.0em}
\end{figure}
%
\begin{equation}
%
f(\bm{x}) = {w_1}p(\bm{x}) +  {w_2}m(\bm{q(x)}), \text{ given } w_1+w_2 = 1, 
\end{equation}
where 
$\bm{x}\in \R^3$ is a  
vector defining the end effector position of the robotic manipulator
and $w_1,w_2$ are scalar weights that determine the importance of the size of the detected object versus the manipulability score. 
Here $m(\bm{q})$
is a metric for dexterity based on the manipulators Jacobian $\bm{J}(q)$ which is a function of the configuration variables $\bm q$ associated with the joints. A standard method in literature for measuring dexterity for a manipulator is to use the determinant of the Jacobian, $\det (\bm{J})$, which is based on the product of its singular values~\cite{park1994kinematic}. The magnitude of the singular values for a Jacobian is proportional to the amount of motion achievable in each dimension of the Cartesian workspace. The dexterity metric (manipulability) we use in this method is defined as
%
%
%
%
%
%
\begin{equation}
m\left(\bm{q}\right) = \sqrt{\det(\bm{J}(\bm{q})\bm{J}(\bm{q})^\mathsf{T})}
\end{equation}	
 
%
\subsection{Gradient Estimation}
In order to compute a gradient $\bm{\nabla f(x)}$ of the objective function the directional derivative, $\nabla_{\bm{v}} f$ can be used. For this work the gradient, $\bm{\nabla f(x)} \in \R^3$, 
is a  vector with components in the principle axes, $x$, $y$ and $z$ of the robots end effector reference frame.  The directional derivative is the derivative in the direction of a unit vector, 
$\bm{v} = \left[ v_x, v_y, v_z \right]^T$, $||\bm{v}||=1$,   
defined as 
\begin{equation}\label{dir_derivative}
%
\nabla_{\bm{v}}  f \equiv \nabla f \cdot \bm{{v}} = \frac{\partial f}{\partial x}{v}_{x} + \frac{\partial f}{\partial y}{v}_{y} +  \frac{\partial f}{\partial z}{v}_{z}.
\end{equation}
This relationship can be used to compute the gradient given measurements of the objective function along different directions. 
Each measurement at different directions can be used to estimate the components of the gradient in the principle axes, $x$, $y$ and $z$.
\begin{figure}
    \centering
    \includegraphics[ width=0.8\columnwidth]{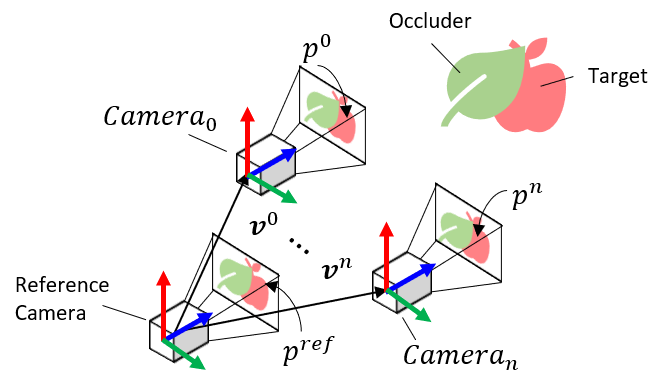}
    \caption{Diagram of the camera array. The camera array contains a reference camera and $n$ cameras translated with respect to the reference camera by the translation vector, $v^n$. Each camera applies the same semantic vision method creating a score of the target object defined as $p^{ref}$ and $p^n$ for the reference camera and surrounding cameras respectively. This enables the camera to determine the next best view based on gradient methods depending on the score from each camera.}
    \label{fig:cam_array}
    \vspace{-1.0em}
\end{figure}

\begin{figure}
    \centering
    
\end{figure}

\begin{figure}[tbp]
    \centering
         \includegraphics[ width=0.8\columnwidth]{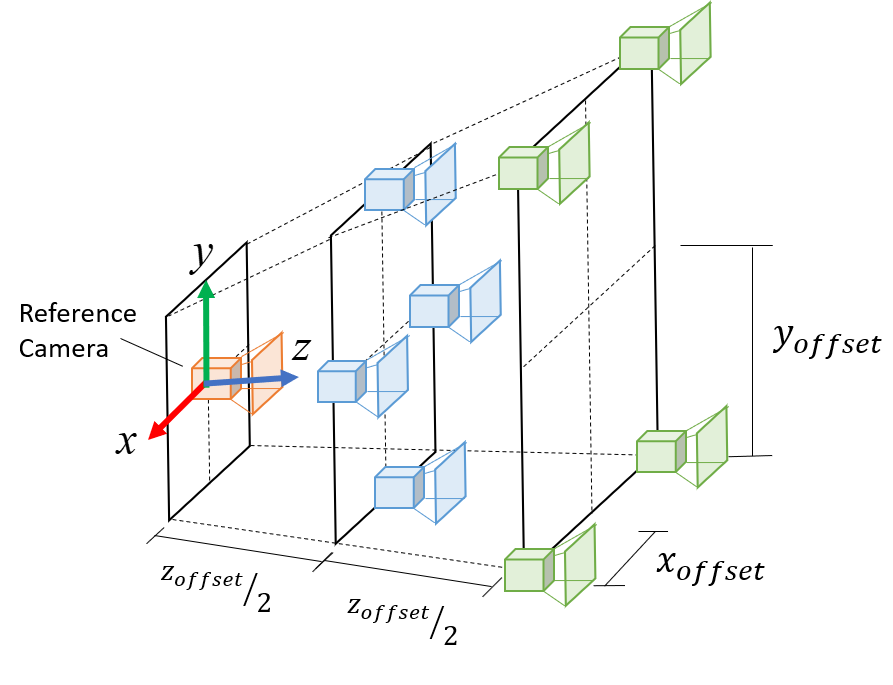}

    %
    %
    \caption{Pi Camera Array. (a) Diagram indicating how the camera array is constructed. (b) Image of the 3D printed Pi Camera array mounted on the robot arm. The camera contains 9 pi cameras and 10 pi zero computer modules.}
    \label{fig:camera_layout}
    \vspace{-1.0em}
\end{figure}

The directional derivative can be approximated by computing the finite difference as the difference between the objective function at the reference camera, $f(\bm{x})$ and a step in the direction of the other cameras. Letting $\bm{\hat{v}^i}$ denote the unit vector in the direction of the $i$th camera
we can write, 
\begin{equation}\label{finite_diff}
\nabla{f} \cdot \bm{\hat{v}^i} \approx \frac{\Delta{f(\bm{x})}}{h} = \frac{f(x+h\bm{\hat{v}^i})-f(x)}{h}
\end{equation}
$h$ is the step size of the finite difference and for this work is the distance between the reference camera and the $i$th camera ($h = \lVert \bm{\hat{v}^i} \rVert$).  

In order to perform gradient ascent with respect to the objective function a least squares approximation can be performed given multiple numerical measurements of the directional derivative for all cameras with respect to the reference camera. This can be achieved by computing the objective function from all cameras and using a first order approximation of the derivative (equation \ref{finite_diff})
\begin{equation}
    \bm{\Delta{f}} = {\matr{V}} \mathbf{\nabla f} \\
    \label{eq_directder}
\end{equation}
where the rows of $\matr{V}\in \R^{n \times 3}$ contain the direction vectors of each 
of the $n$ cameras, with respect to the central view, 
\begin{equation}
\matr{V} = \begin{bmatrix} \bm{v}^0 & \hdots & \bm{v}^n\end{bmatrix}^T
\end{equation}
and $\bm{\Delta{f}}\in \R^n$ is a  vector of finite differences approximations of the directional derivative computed between the reference camera and each camera in the array defined as 
\begin{equation}
    \bm{\Delta{f}} = \begin{bmatrix} \Delta{f_0(\bm{x})} & \hdots & \Delta{f_n(\bm{x})}\end{bmatrix}^T
\end{equation}
The gradient can then be computed by solving \eqref{eq_directder} for $\bm{\nabla {f}}$ in a least-squares sense. This can be written in closed form as  
\begin{equation}\label{least_squares}
 \bm{\nabla {f^*}}=(\matr{V}^T\matr{V})^{-1}\matr{V}^T\bm{\Delta{f}}
\end{equation}
Where $\bm{\nabla {f^*}}$ denotes an estimatation of the true gradient. Once the gradient is estimated the next step is to use gradient ascent to move in the direction that maximises the objective function. This direction is taken in the frame of the end effector,  which is also the frame of the reference camera. The next position of the end effector, $\bm{x}_{k+1}$ is computed using
\begin{equation}
    \bm{x}_{k+1} = \bm{x}_k + \alpha \bm{\nabla f_k(x_k)}
\end{equation}
For this work, the gradient ascent method is applied to the translational component of the end effector, $\bm{x}_k$. This was chosen as the orientation of the next best view for the end effector can be computed in a simplified manner. The stopping condition was chosen as $\frac{1}{m}\sum_{k=1}^{m} \lVert {\nabla f_k}\rVert > \epsilon$ in order to account for the potential noise in the estimate of the gradient. A maximum target score, $p^{max}$, was also added as a stopping condition which determined the stopping distance between the camera and the target. 

An assumption is made that a change in orientation (no translation) of the camera has little to no effect on the score of the target, unless the target is at the edge of the cameras field of view. Therefore, for this work we compute the roll, $\phi$ and pitch, $\theta$ angles required to ensure the centre of the target is in the centre of the image, where yaw (rotation about the cameras $z$ axis) is ignored. The roll and pitch required to centre the target are calculated as the normalised distance (along both the u and v axis) between the centre of the segmented target and centre of the image and multiplied by the camera FOV.  
%
%
%
%
The proposed method described is referred to as 3D Move to See (3DMTS) and is summarised in Algorithm~\ref{alg:move_to_see}.

\renewcommand{\algorithmicrequire}{\textbf{Input:}}
\begin{algorithm}
    \caption{3D Move to See (3DMTS)}\label{move_to_see_alg}
    \label{alg:move_to_see}
    \begin{algorithmic}[1]
        \Require $\epsilon$, $p^{ref}$ and $p^{max}$ 
        \State $\matr{V} \Leftarrow \begin{bmatrix} \bm{v}^0 & \hdots & \bm{v}^n\end{bmatrix}$ \Comment{Initialise direction vector matrix }
        \While{ Stopping Criteria $> \epsilon$ \textbf{and} $p^{ref} < p^{max}$}
            \For{each camera in array}
                \State Apply image segmentation to compute $p$
                \State Calculate $q$ using inverse kinematics
                \State Compute manipulability, $m(q)$ 
            \EndFor
            \State Compute gradient $\bm{\nabla f_k(x_k)}$
            \State Compute roll and pitch angles for reference camera
            \State Move end effector to $x_{k+1}$, $\phi_{k+1}$ and $\theta_{k+1}$
        \EndWhile
    \end{algorithmic}
\end{algorithm}

\subsection{Image-Based Segmentation}
Different semantic segmentation methods can be used within this method. We use the same segmentation approach as~\cite{Lehnert16_1:conference, lehnert2017} which consists of colour segmentation, as such we apply this approach to better view red sweet pepper.

The method consists of training a multi-variate Gaussian to model the distribution of the target object,

\begingroup
{\footnotesize
\begin{equation} \label{eq:multi_var_gaussian}
 p\left(\vecx \mid \theta \right) = \left(2\pi \right)^{-\frac{D}{2}} \left| \mySigma \right|^{-\frac{1}{2}} 
                                          \exp\left[ -\frac{1}{2}\left(\vecx-\mean\right)^T \mySigma^{-1}\left(\vecx-\mean\right) \right],
\end{equation}}
\endgroup

\noindent where $\mean$ is the mean of the data, $\mySigma$ is the covariance matrix (assumed to be diagonal), and $D$ is the number of dimensions of the feature vector $\vecx$; we use rotated HSV values as features $D=3$ described below.
To determine if a pixel value corresponds to the object a threshold $\tau$ is applied to Equation~\ref{eq:multi_var_gaussian}.

As with the prior work of Lehnert et al.~\cite{lehnert2017} we use a rotated HSV colour space. This is obtained by first transforming the red, green, blue (RGB) colour representation to the hue, saturation and value (HSV) colour space. 
The angular value for the hue is rotated by 90$^{\circ}$ so that distances as the red value ranges from from approximately 315$^{\circ}$ to 45$^{\circ}$ (centred around 0), this is a rotated HSV colour space.

\subsection{Baseline Method}
In order to analyse the performance of the proposed method, a simple naive approach is used as a baseline. This approach uses a single RGB-D camera to estimate the target location using depth from a single perspective. The same image segmentation is applied to the RGB image and the depth of the target centre is used to move the camera in a linear trajectory towards the target. This method would usually work for cases with no occlusions but may struggle when occlusions are present and therefore is a good baseline for the proposed move to see approach.

\section{Experiments and Results}

The proposed method was firstly implemented within a simulation environment in order to analyse the effects of different parameters, such as the distance between multiple perspectives in the array and weights in the objective function. The proposed method was then tested on a real robotic platform developed for autonomous harvesting~\cite{lehnert2017}. 

\subsection{Experiment 1 - Simulation}

The first experiment investigates the proposed method within a realistic simulation of a robotic platform and a range of single target and occluder scenarios. The simulator used within this work was VREP~\cite{rohmer2013v}.

A set of trials were run from an initial robot configuration in which a range of parameters where modified. This included moving the position of an occluding object in a 2D grid in front of the target while varying the occlusion angle to test different scenarios.

Table~\ref{tbl:parameters} lists the parameters that were modified, which includes: translation of the target and occlusion in world coordinates, offset parameters of the camera array and relative weights in the objective function. 

\begingroup 
\setlength{\tabcolsep}{4.5pt}
\begin{ctable}[%
 caption = {Parameters for Sim and Real Robot Experiments},
 label = tbl:parameters,
 width = \columnwidth,
 doinside = \footnotesize,
 pos = tp,
 ]{ccc}{}
 {\FL \textbf{Parameter} & \textbf{Sim - Range/Value} & \textbf{Real - Value} \ML 
target position $(y)$ & [-0.1, 0, 0.1]  & N/A\\
target position $(z)$ & [-0.1, 0, 0.1]  & N/A \\
occlusion position $(y)$ & [-0.1, 0, 0.1] &  N/A \\
occlusion position $(z)$ & [-0.1, 0, 0.1] &  N/A \\
occlusion angle $\theta$ & [-45, 0, 45]  &  [0, $\pm45$, $\pm90$, $\pm135$, 180]\\
$w_1$ & [1.0 ,0.8] & 1.0\\
$w_2$ & [0.0, 0.2] & 0.0\\
camera array radius $r$ & [0.06, 0.09. 0.12] & 0.048 \\
pixel noise $\sigma$ & [0.001, 0.01] & N/A \\ 
step size, $\alpha$ & 0.001 & 0.001 \\ 
Stop condition [$\epsilon$, $p_{max}$] & [1.5, 0.4] & [1.5, 0.3] \LL 
 }
\end{ctable}
\endgroup

\begin{figure}[tbp]
    \centering
    \begin{subfigure}[]{0.33\columnwidth}
        \includegraphics[ width=\textwidth]{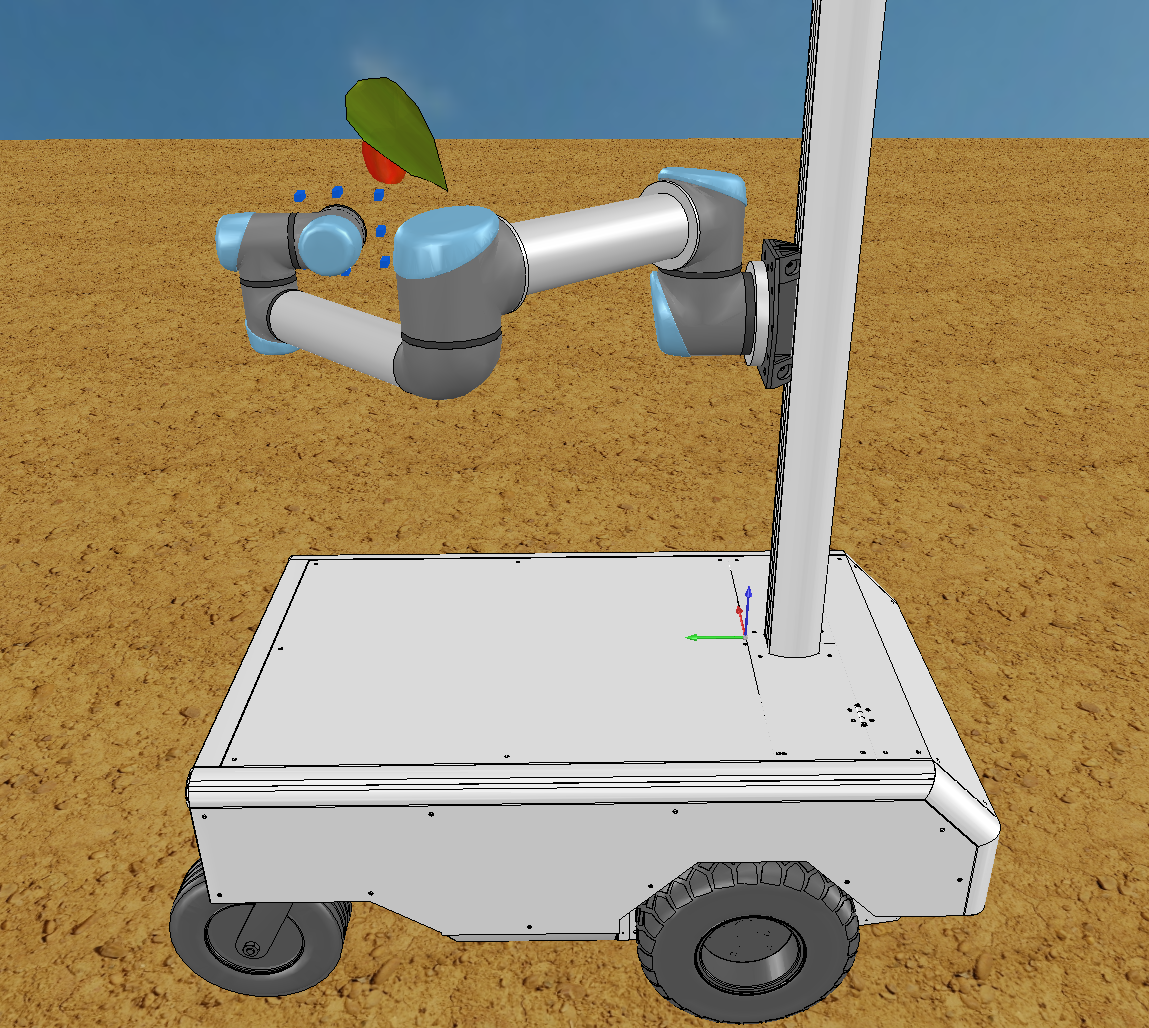}
        \caption{}
    \end{subfigure}
    \begin{subfigure}[]{0.31\columnwidth}
         \includegraphics[ width=\textwidth]{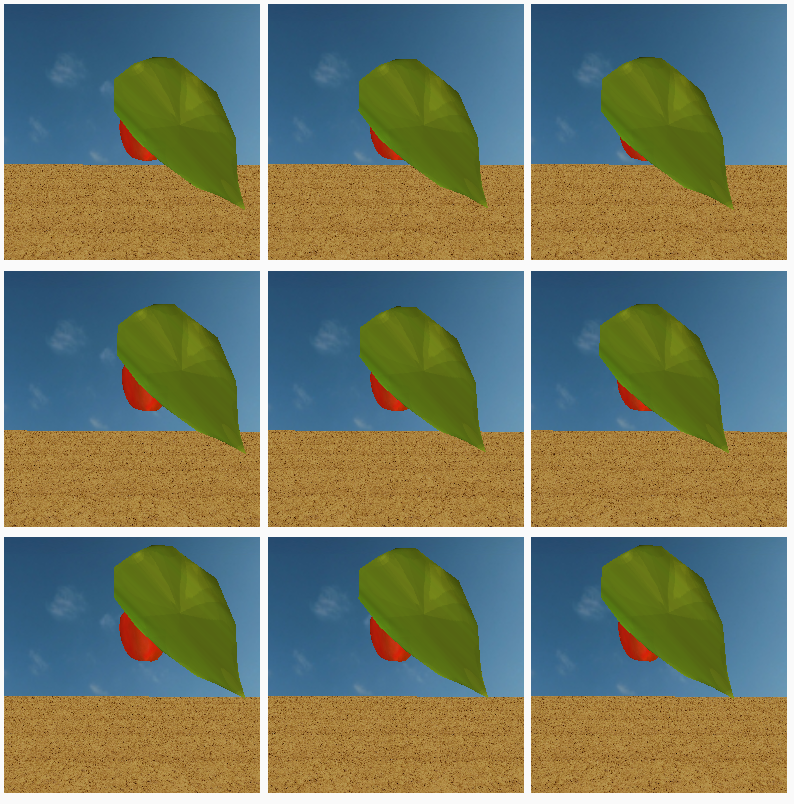}
        \caption{}
    \end{subfigure}
    \begin{subfigure}[]{0.31\columnwidth}
         \includegraphics[ width=\textwidth]{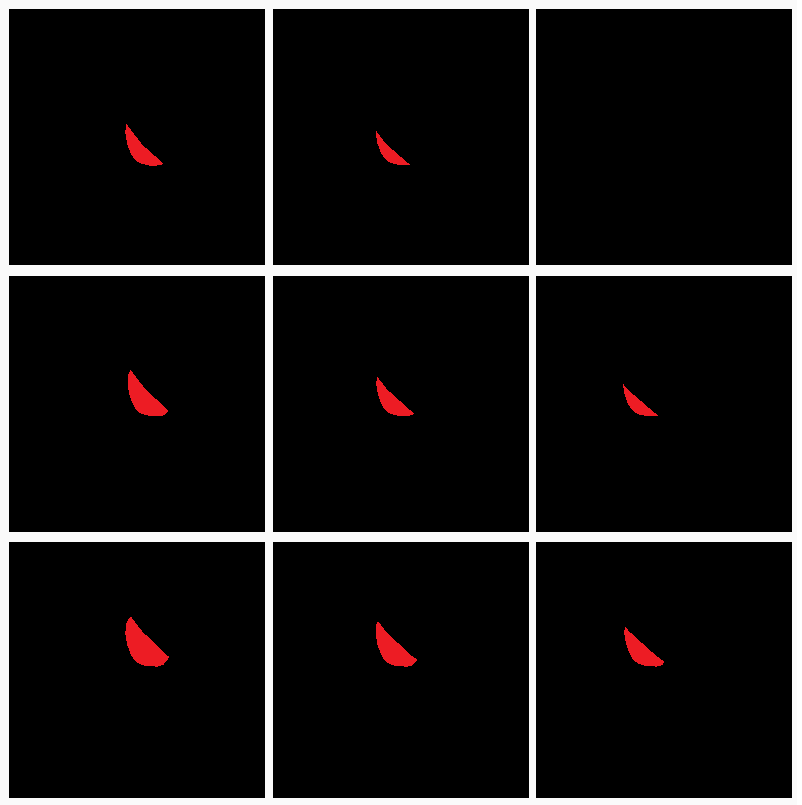}
        \caption{}
    \end{subfigure}
    \caption{(a) Images of the simulation environment including the example 7DoF robot platform and example single target and single occlusion. Example images taken from the simulated camera array from multiple perspectives. (b) RGB images before the semantic segmentation is applied. (c) Images after the semantic segmentation is applied.}
    \label{fig:simulation}
    \vspace{-1.0em}
\end{figure}

Table~\ref{tbl:objective} compares the value of the objective function $f$ at the final step $N$ between the naive method and our proposed method. We calculate $\bar{f_N}$, the mean of $f_N$ over 400 experiments with varying target and occlusion positions. We also calculate the increase in target image area (from segmentation) from the start to the end of the trajectory $\Delta A$ for both the baseline and our proposed methods. We anticipate that increased target area will improve the detection of key features for grasping and harvesting.

\begin{ctable}[%
 caption = {Comparison of Methods in Final Objective Function and Area Increase from Start to End Images},
 label = tbl:objective,
 doinside = \footnotesize,
 pos = tp,
 ]{cc | cc |cc}{}
 {\FL \textbf{Parameter} & \textbf{Value} & \multicolumn{2}{c}{\textbf{Naive}} & \multicolumn{2}{c}{\textbf{Proposed}} \\
 & & $\bar{f}_N$ & $\Delta A$ [\%] & $\bar{f}_N$ &$\Delta A$ [\%] \ML
$\theta$ & 0 [$^\circ$] & 0.128 & 10.5 & \textbf{0.234} & \textbf{21.3}\\
$\theta$ & 45 [$^\circ$] & 0.101 & 8.9 & \textbf{0.185} & \textbf{19.0}\\
$\theta$ & -45 [$^\circ$] & 0.115 & 9.7 &  \textbf{0.203}&\textbf{17.5} \\
($w_1$,$w_2$)& [1,0] & 0.113 & 9.8 & \textbf{0.205} & \textbf{22.4}\\
($w_1$,$w_2$)& [0.8,0.2] & 0.137 & 9.6 & \textbf{0.232} & \textbf{19.0}\\
$r$ & 0.06 [m] & 0.118 &  9.7 &\textbf{0.284}& \textbf{26.9}\\
$r$ & 0.09 [m] & 0.113 & 9.7 & \textbf{0.186} & \textbf{17.0}\\
$r$ & 0.12 [m] & 0.133 & 9.7 & \textbf{0.133} & \textbf{11.8} \\
$\sigma$ & 0.001 & 0.115 & 9.7 & \textbf{0.240}& \textbf{22.6}\\
$\sigma$ & 0.01 & 0.115 & 9.7 &\textbf{0.176} &\textbf{16.0} \LL 
 }
\end{ctable}

\begin{figure}[tbp]
\vspace{-1.0em}
    \centering
    \begin{subfigure}[]{0.48\columnwidth}
        \includegraphics[ width=\textwidth]{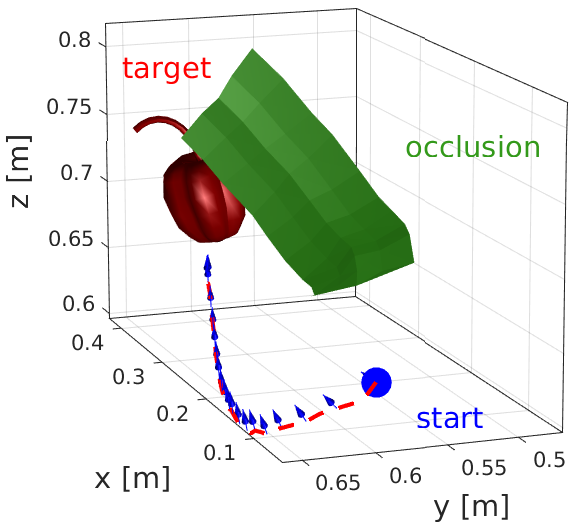}
        \caption{}
    \end{subfigure}
    \hfill
    \begin{subfigure}[]{0.48\columnwidth}
        \includegraphics[width=\textwidth]{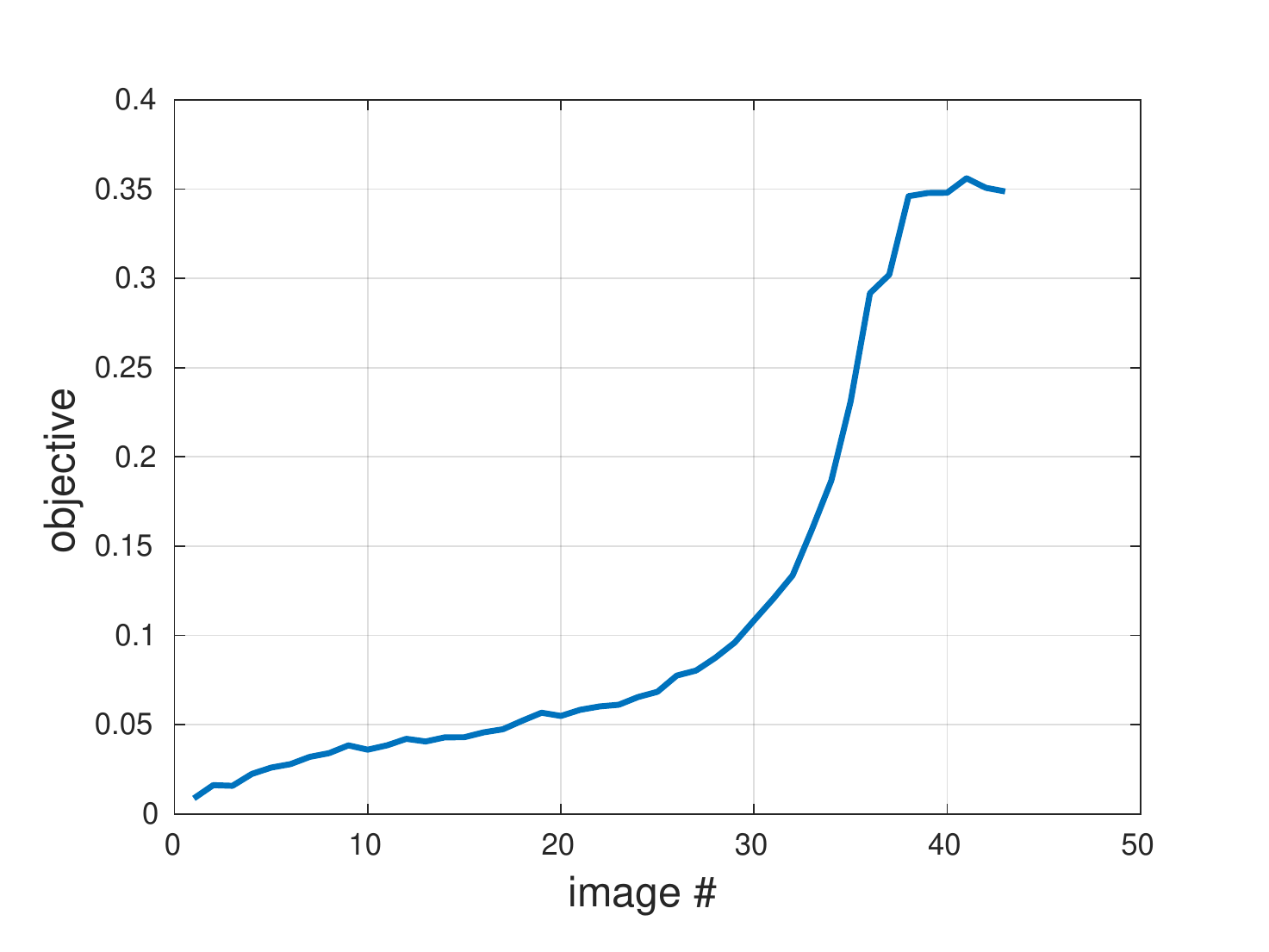}
        \caption{}
    \end{subfigure}
    \caption{Sample trial of the robotic arm in simulation where (a) is the trajectory and (b) is the objective function. The trajectories are generated using 3DMTS methodology proposed. The red line represents the end effector position and the blue arrows indicate the orientation.}
    \label{fig:sample}
    \vspace{-1.0em}
\end{figure}

\begin{figure}[tbp]
    \centering
    \includegraphics[width=0.8\columnwidth]{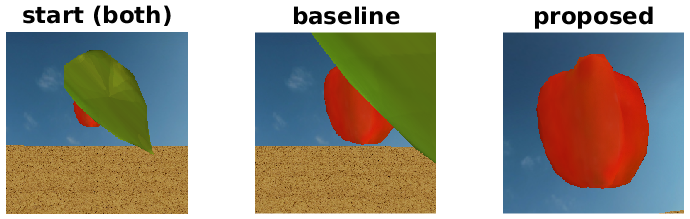}
    \caption{Comparison of final images. With the baseline method, the target is still partially occluded, while with the proposed method, the target is completely revealed.}
    \label{fig:endImages}
\end{figure}

\subsection{Experiment 2 - Real Robot Platform}

The proposed move to see method was tested on a real robotic platform that was designed for autonomous harvesting of sweet peppers~\cite{lehnert2017}. Harvesting is a challenging problem where the target object naturally varies in shape and the scene is unstructured and occluded by leaves. The harvesting platform which includes a 7DoF robot manipulator was retrofitted with a prototype 3D camera array end effector. The prototype 3D camera array was created using nine raspberry pi cameras and ten raspberry pi zero computer modules (See Fig.~\ref{fig:FrontImage}b). The experiment setup can be seen in Fig.~\ref{fig:FrontImage}a.

The experiment was conducted in a similar manner as the previous simulation experiment, where a single target (plastic sweet pepper) was placed in an unstructured scene (fake plant) and tested for a range of different occlusion scenarios. The different occlusions where parameterised by an angle of rotation. The 3D camera array had fixed offsets which approximated to a radius of 0.048 m based on fixed offsets of $xyz_{offset} = [0.027, 0.027, 0.03]$ (see Fig.~\ref{fig:camera_layout}). The parameters of the setup for the real robot experiment can be found in Table~\ref{tbl:parameters}. 
As an initial experiment, the robot configuration was chosen such that the condition of the $\bm{J}$ did not significantly affect the manipulability score. 

\begin{figure}[tbp]
    \centering
    %
   	\begin{subfigure}[]{0.48\columnwidth}
        \includegraphics[width=\textwidth]{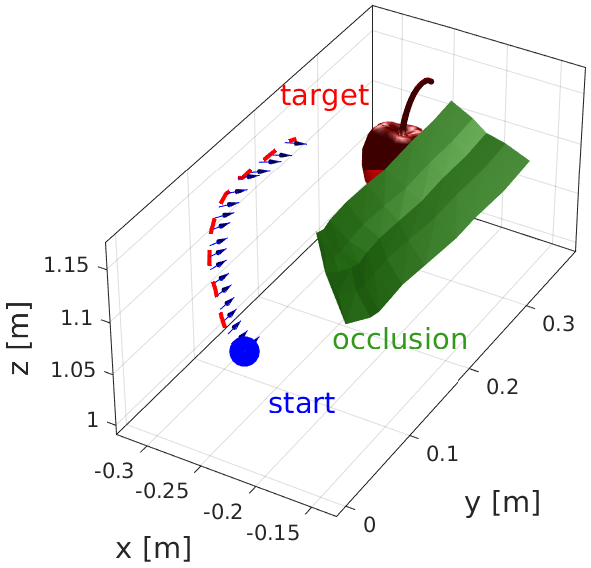}
        \caption{}
    \end{subfigure}
    \hfill
    \begin{subfigure}[]{0.48\columnwidth}
        \includegraphics[width=\textwidth]{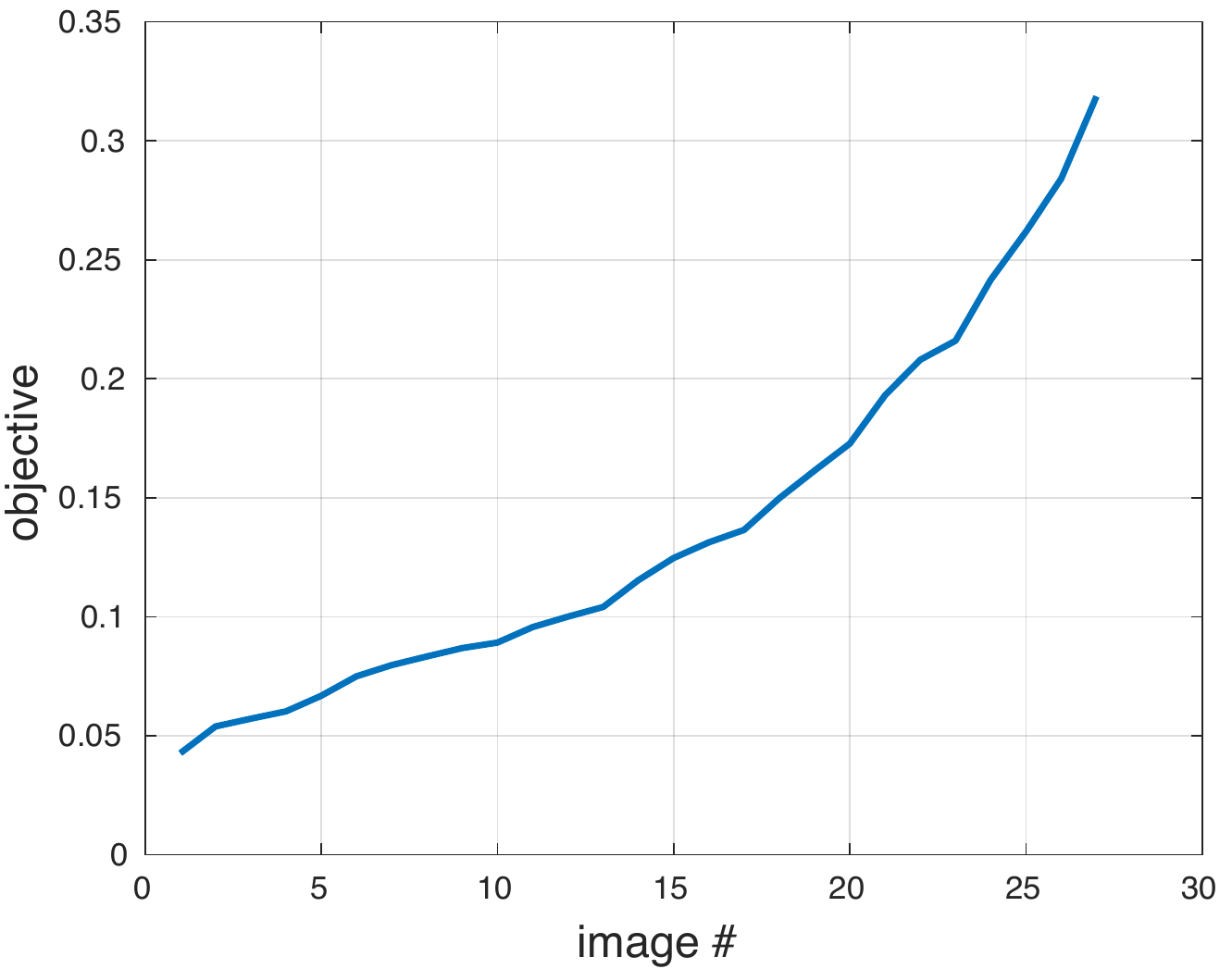}
        \caption{}
    \end{subfigure}
    \caption{(a) Sample trajectory of the robotic arm navigating around the occlusion. The target and occlusion are shown for visualisation. (b) The corresponding objective function.}
    \label{fig:real}
\end{figure}

The move to see method was executed until the stopping condition was met. The starting and ending image and objective function along with the end effector trajectory of the robot manipulator were recorded. Fig.~\ref{fig:real} shows the results for a single trial of one scenario. The baseline method was also implemented using an RGB-D camera (Intel Realsense SR300) and the same results were recorded for the occlusion scenarios.

A measure of the performance for the experiment is the percentage increase of the target size between the start and end image. This indicates whether the method finds the locally optimal view given the starting condition.  
The mean objective function values and percentage increase for different occlusion angles are presented in Table~\ref{tbl:objectiveReal}. The overall performance of the proposed method obtained a mean increase in target size by $29.3\%$ whereas the baseline method obtained a mean increase in target size by $9.17\%$.

\begin{figure}[tbp]
    \centering
    \includegraphics[ width=0.8\columnwidth]{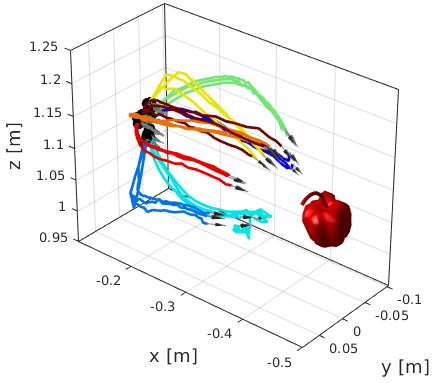}
    \caption{Overlay of robot trajectories for different occlusion angles by colour that all converge towards the target sweet pepper. The occlusions are not shown.}
    \label{fig:octopus}
    \vspace{-1.0em}
\end{figure}

\begin{ctable}[%
 caption = {Comparison of Methods for Varying Occlusion Angle with Robot Arm},
 label = tbl:objectiveReal,
 doinside = \footnotesize,
 pos = tp,
]{c | cc | cc}{}
 {\FL \textbf{Occlusion} &  \multicolumn{2}{c}{\textbf{Naive }} & \multicolumn{2}{|c}{\textbf{Proposed}} \\
 $\theta$ [$^\circ$] & $\bar{f}_N$ & $\Delta A$ [\%] & $\bar{f}_N$ & $\Delta A$ [\%] \ML
0       & \textbf{0.41}  & \textbf{38.6}  & 0.313         & 28.9\\
45      & 0.00     & -5.4  & \textbf{0.30} &\textbf{29.1} \\
90      & 0.13  & 11.1  & \textbf{0.31} & \textbf{29.8}\\
135     & 0.04  & 1.9   & \textbf{0.35} & \textbf{31.8}\\
180     & 0.01  & -2.0  & \textbf{0.34} & \textbf{30.5}\\
-135    & 0.03  & -3.5  & \textbf{0.31}  & \textbf{28.3}\\
-90     & \textbf{0.42}  & \textbf{37.7}  & 0.32 & 27.9\\
-45     & 0.00     & -5.1  & \textbf{0.32}  & \textbf{28.3}\LL 
 }
\end{ctable}

\begin{figure}[t]
\centering
\scriptsize
\setlength\tabcolsep{1.5pt} 
\vspace{1em}
\begin{tabular}{cccc}
Baseline Start	& Baseline End & Proposed Start & Proposed End \\
\includegraphics[width=0.24\columnwidth]{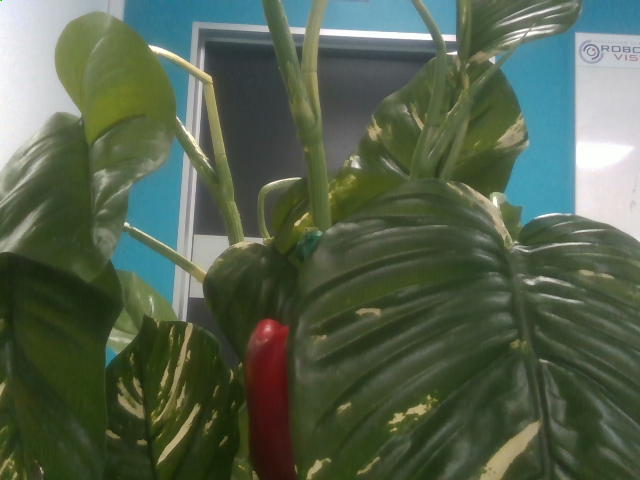}
& 
\includegraphics[width=0.24\columnwidth]{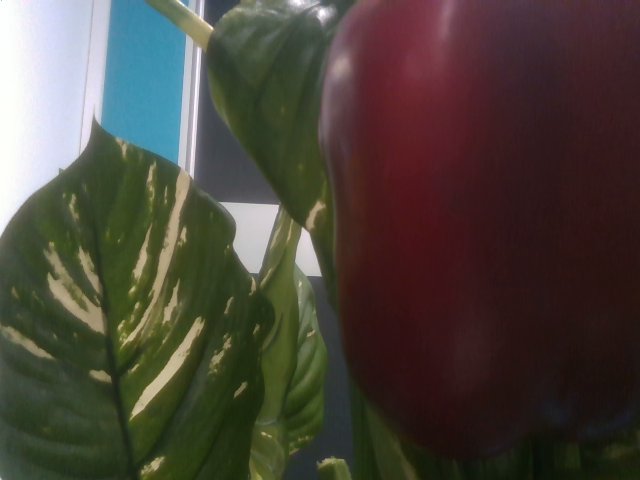}
&
\includegraphics[ width=0.24\columnwidth]{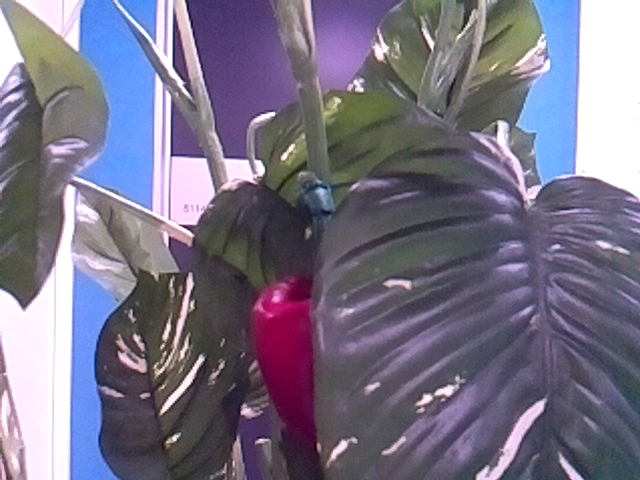}	
&
\includegraphics[ width=0.24\columnwidth]{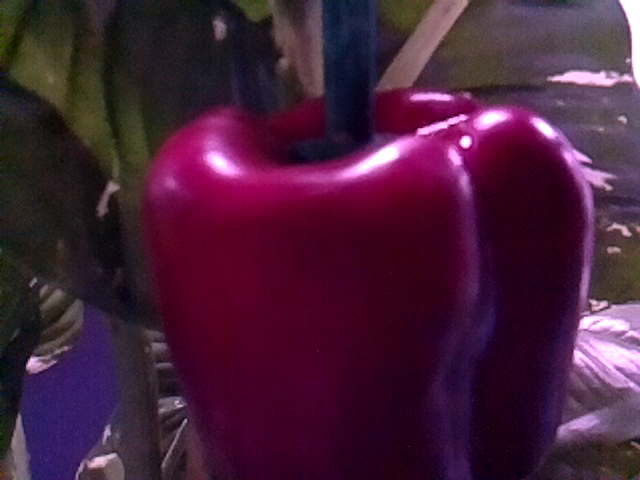}	
\\
\includegraphics[ width=0.24\columnwidth]{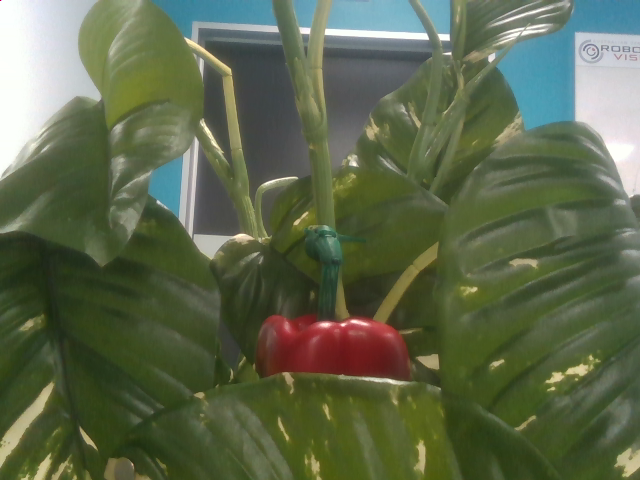}	
&
\includegraphics[width=0.24\columnwidth]{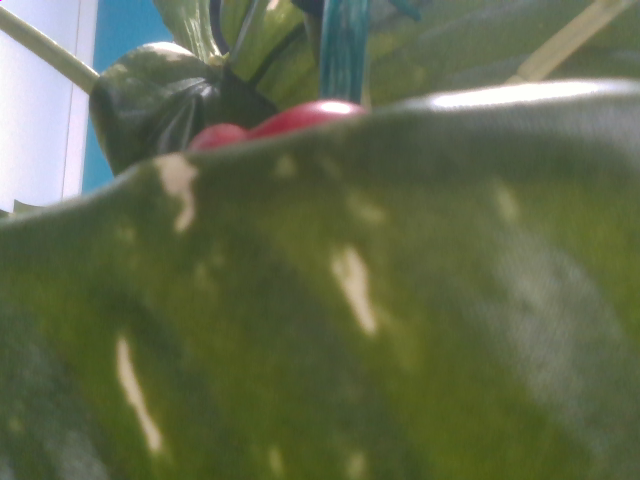}	
&
\includegraphics[width=0.24\columnwidth]{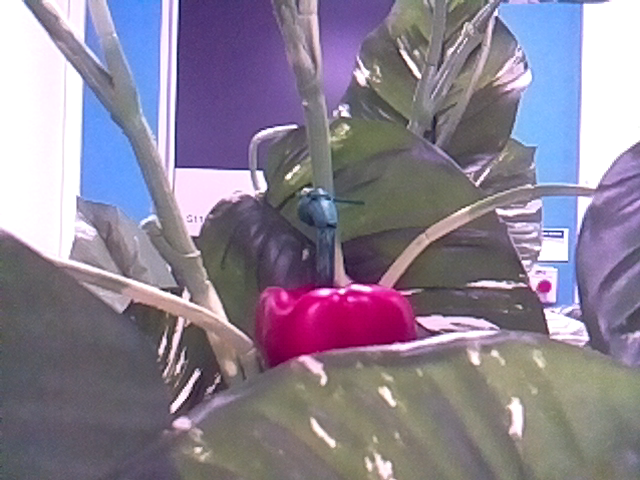}	
&
\includegraphics[width=0.24\columnwidth]{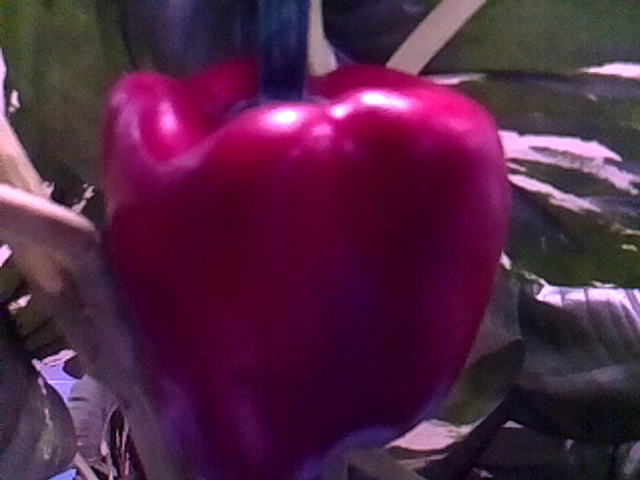}	
\\
\includegraphics[width=0.24\columnwidth]{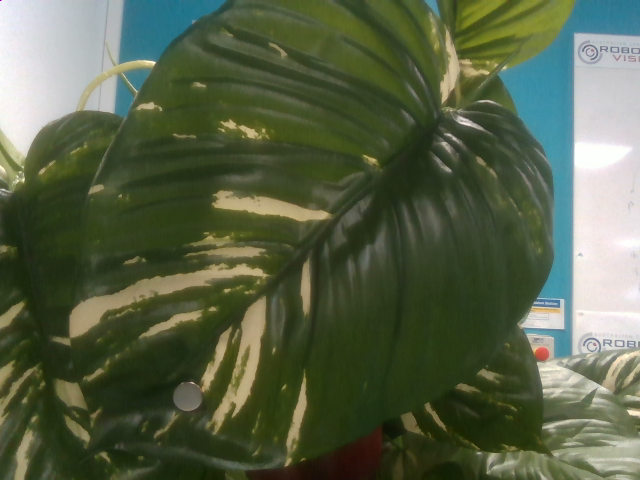}	
&
\includegraphics[ width=0.24\columnwidth]{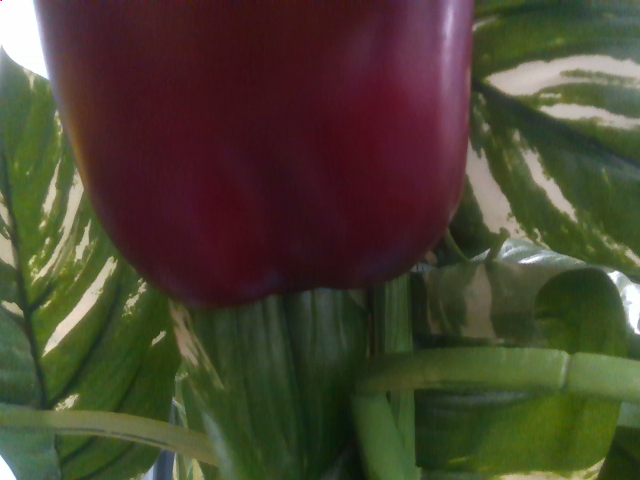}	
&
\includegraphics[ width=0.24\columnwidth]{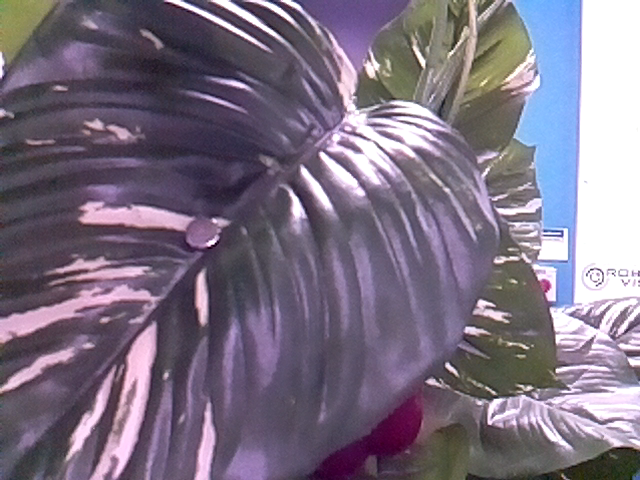}	
&
\includegraphics[width=0.24\columnwidth]{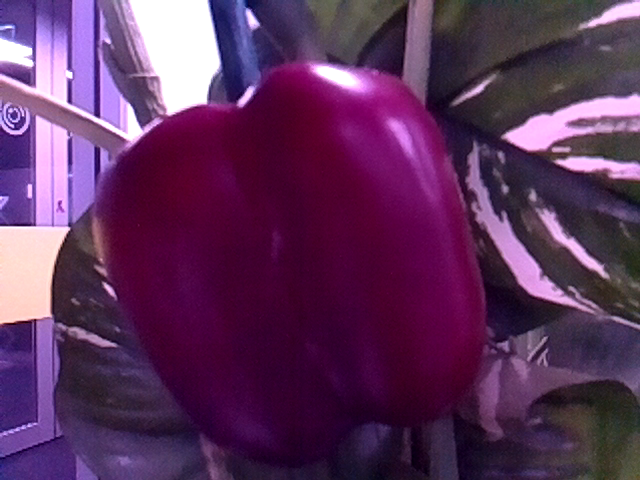}
\\
\end{tabular}
\caption{Comparison of baseline method to the proposed initial and final views, respectively. The baseline method does not always navigate the occlusions, while the final images from our proposed method provide a much more centred view, and are much more useful in applications such as harvesting the sweet pepper.
}
\vspace{-1.5em}
\label{fig:comparison}
\end{figure}
\section{Discussion and Conclusion}

This paper presents the 3D Move To See (3DMTS) method for the next best view problem using a 3D multiple perspective camera array and a gradient ascent approach over an objective function. Howeve, the proposed 3DMTS method has some limitations. 

Firstly, the method relies on the performance of the semantic segmentation and can cause inaccuracies in estimating the gradient if false positives or negatives occur. However, the 3DMTS can recover from inaccuracies as it samples the gradient at each time step effectively smoothing over the errors. Another limitation is that the trajectory of the end effector is dependent on the surface of the objective function and is hard to predict. Other criteria in the objective could be used to shape the trajectory. Other limitations include the requirement to have multiple cameras (nine for this work) including processing each image. Future work could look at methods to reduce the number of cameras by possibly estimating the gradient from a single perspective using the 3D camera array as a method for capturing training data.

Future work will look into the effects on the final view for different initial configurations (including challenging configurations near singularities) of the robotic manipulator. Other future work would look at further optimising the parameters of the method such as the camera array size. Improvements to the hardware is also planned such as improving the bandwidth of the camera to improve the visual servoing rate. As this is a visual servoing method future work could also look at dynamic environments where the target or scene is moving. The objective function approach to 3DMTS is versatile, and can be customised for other applications. For example, a grasping metric could be used to adapt the system to find locally optimal grasp poses for manipulation. 

The primary motivation for this method is for finding the next best view in order to handle occlusions of objects in unstructured environments. It was shown within simulation and on a real robotic platform that moving the end effector using the gradient of an objective function leads to a locally optimal view of the object of interest, even amongst highly occluded scenes and unstructured environments. The experiments demonstrated that the proposed 3DMTS method performed better in all except a few scenarios with a three fold increase in target size compared to the baseline method. The increased target size will improve the view of the object of interest for further grasping and manipulation, ultimately leading to improved robot harvesting accuracy.

\vspace{-2mm}
\bibliographystyle{IEEEtran}
\bibliography{./bibs/mccool_bibs,./bibs/ICRA2016_Refs,./bibs/dtsai_icra2019,./bibs/vision_ag}



%


%

\end{document}